**Title:** trACT: temporal revelation Airborne Camera Trap

**Authors:**
Oliver Bimber[1*], Rakesh John Amala Arokia Nathan[1], Mohamed Youssef[1], Vinayak Lal Bhatnagar[1], Ralf Berger[2], Klaus Hackländer[3,4]

**Affiliations:**
[1]Department of Computer Science, Johannes Kepler University Linz, Austria
[2]Institute of Space Research, German Aerospace Center, Berlin, Germany
[3]BOKU University, Institute of Wildlife Biology and Game Management, , Vienna, Austria
[4]German Wildlife Foundation (Deutsche Wildtier Stiftung), Hamburg, Germany

*Corresponding author. Email: oliver.bimber@jku.at

**Abstract:** Effective remote monitoring and surveillance using drones are frequently impeded by severe environmental and thermal clutter, dynamic vegetation, target camouflage, and system latency. Drawing inspiration from the hunting strategies of birds of prey that hover and stabilize their vision to isolate subtle ground motion, we introduce *trACT* (temporal revelation Airborne Camera Trap), a lightweight, real-time aerial robotics framework designed for autonomous consumer drones. The system integrates Temporal Max Pooling (TMP), a low-level signal processing method that transforms imperceptible movement across a rolling integration window into robust value and time encodings, with self-supervised motion anomaly detection to isolate target motion from background environmental motion caused by wind gusts and drone drift. To overcome mechanical and processing delays, *trACT* combines motion prediction with automated gimbal-stabilized optical zoom verification and equitable multi-target verification balancing. Extensive real-world field experiments in densely forested wildlife habitats and surveillance scenarios demonstrate that *trACT* successfully bridges the gap between wide-area aerial monitoring and precise, autonomous target verification under challenging operational conditions.

**One Sentence Summary:** Inspired by birds of prey, *trACT* is an airborne camera trap designed for autonomous target detection and monitoring in cluttered environments.

**Main Text:**

## INTRODUCTION

Sustainable wildlife management, including conservation, control and use, relies on the effective monitoring of animal populations, behaviors, and habitat conditions, as well as the implementation of robust anti-poaching measures. Traditional approaches depend heavily on manual field surveys and patrolling, which are labor-intensive, costly, and often limited in spatial and temporal coverage. To overcome these limitations, camera traps (1-3) and camera drones (4-11) have emerged as highly effective vision sensors for large-scale wildlife monitoring, offering the ability to survey extensive areas while minimizing disturbance to wildlife.

Recent advances in signal processing and deep learning have enabled the automatic detection and counting from aerial imagery using drones. Several studies have employed object detection frameworks to identify and monitor various species, including mammals and birds, achieving accurate population estimation over large geographical regions (12-20). Aerial synthetic aperture imaging can mitigate vegetation-induced occlusion for detection (20,21), provided that the approximate locations of the animals are known and they remain stationary during the scanning process. These approaches have significantly reduced the manual effort required for wildlife surveys while improving monitoring efficiency. In addition to wildlife monitoring, these aerial perception systems play a critical role in anti-poaching operations (22-24), where detection of human intruders, vehicles, or suspicious activities enables rapid response and enhanced protection of endangered species in large and often inaccessible conservation areas.

Beyond spatial-based detection and counting, extracting motion information from video data remains fundamental to studying wildlife behavioral ecology, migration patterns, and population dynamics (1-3, 25-28). Existing computer vision approaches include absolute frame differencing (27), optical flow and background modeling (25), spatiotemporal segmentation (1), and deep neural networks (2,3,28). Although these methods demonstrate promising results under relatively simple, near-field conditions, achieving real-time performance often requires compromising spatial resolution or increasing computational overhead. Furthermore, they are frequently unsuitable for far-field (high-altitude aerial) monitoring, suffering significant performance degradation when targets and their motion occupy only a few pixels or are obscured by occlusion, camouflage, and thermal clutter.

Beyond wildlife management, vision-based surveillance applications suffer from similar problems. They encompasses aerial small-target monitoring (29-35), counter-drone defense (36-40), traffic surveillance (41-44), maritime safety (45-50), border security (51, 52), perimeter and railway intrusion detection (53-56), search and rescue operations (57-59) and security anomaly detection (60). These systems are deployed on unmanned aerial vehicles (UAVs) or on fixed, trackside and pan-tilt cameras. Most approaches perform detection on a frame-by-frame basis, only a few leverage inter-frame motion under platform ego-motion (32-35, 40, 55, 58, 60) or actively reposition to verify detected candidates (44, 48). Performance typically degrades in the far field due to challenges such as few-pixel targets, cluttered and dynamic backgrounds, occlusion, platform jitter, motion blur and the trade-off between real-time processing and resolution on constrained hardware.

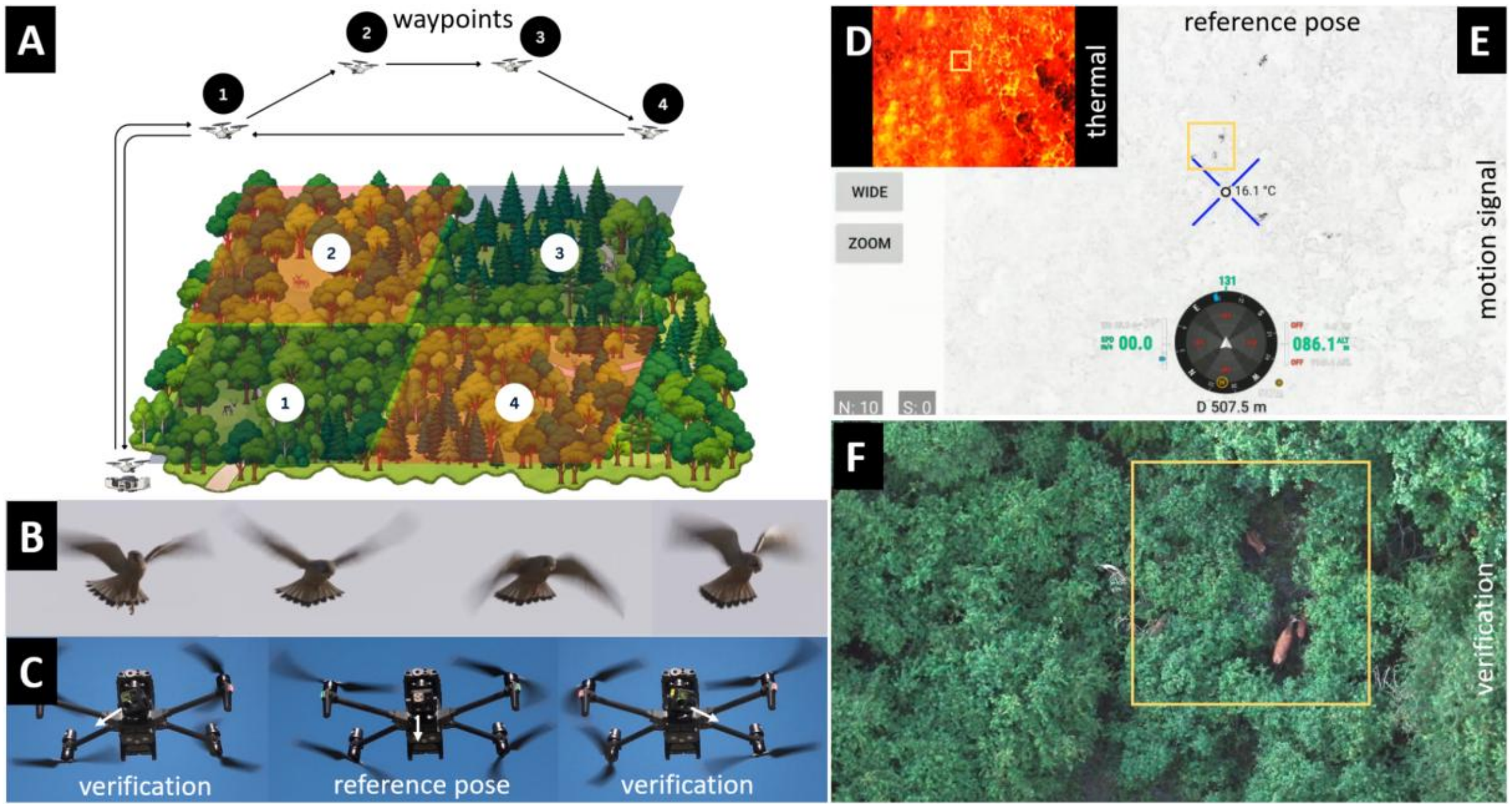

**Fig. 1. Airborne camera trap principle.** Automatic waypoint flights with a camera drone enable the flexible scanning of large regions from high altitudes, while docking stations support continuous charging and remote monitoring **(A)**. Motion detection from higher altitudes draws inspiration from the fixation strategies of birds of prey, such as common kestrels (*Falco tinnunculus*), which hover while stabilizing their heads to detect the motion of ground-level prey **(B)**. Our drone mimics this behavior to detect subtle motion signals from a reference pose across a sliding time window, using wide-angle RGB or thermal images. Once motion is detected, the camera gimbal rotates to center and zoom in on the target. The resulting magnified image (daylight or night-vision RGB) is then used for verification. This scanning process runs directly on the drone hardware at a frequency of approximately 1Hz **(C)**. Wide-angle thermal images recorded from higher altitudes (86 m AGL in this example) obscure moving red deer within thermal clutter and limited spatial detail **(D)**. Temporal Max Pooling (TMP) highlights the animals' motion trails with dark pixels **(E)**. Automatic centering and zooming on the detected targets enable optical verification even from high altitudes (**F**). Boxes highlight the same ground area. See Supplementary Movie S1.

To address these problems, we propose the *temporal revelation Airborne Camera Trap*, *trACT*, a system designed for immediate deployment on consumer drones. Unlike conventional camera traps, *trACT* enables aerial, large-area coverage across flexible observation sites and provides continuous monitoring via automated waypoint flights (Fig. 1A). Operating such a system presents significant challenges: extreme visual clutter from dense vegetation, thermal environmental noise, and motion artifacts caused by wind and drone drift. Additionally, achieving reliable detection from higher altitudes is difficult due to constrained spatial and temporal resolutions, as well as limited onboard processing power.

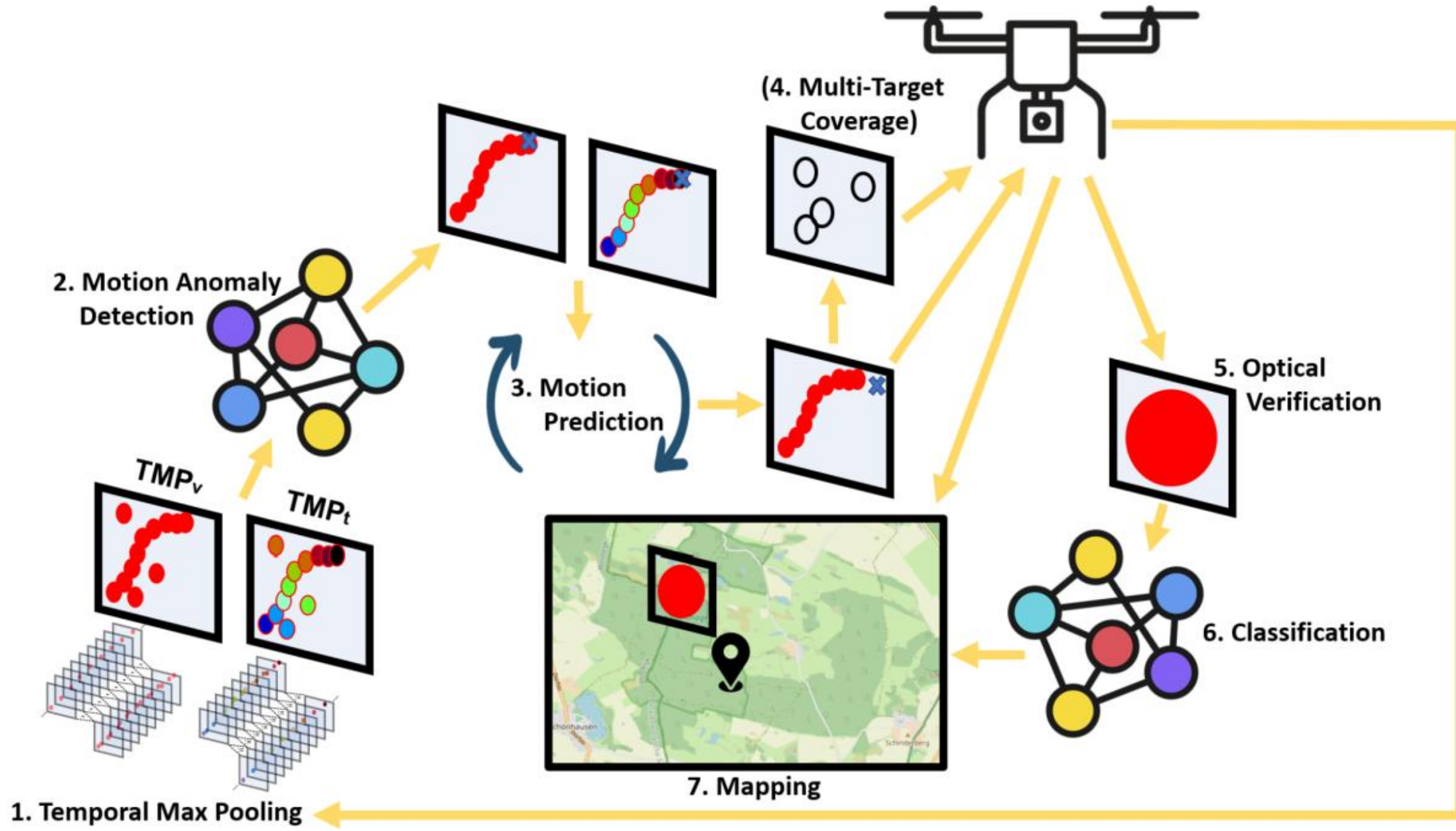

**Fig. 2. The *trACT* processing pipeline.** Temporal Max Pooling (TMP) **(1.)** serves as the initial signal processing component, detecting motion within a rolling image integration window to yield value and time encodings of motion trails. These channels are used for the self-supervised training and inference of a motion anomaly detection network **(2.)**, which distinguishes targets from common background motion caused by wind (such as swaying vegetation and drone drift). Combined with Kalman filtering **(3.)**, the TMP time channel enables the prediction of the target's actual position once processing latency has elapsed. An optional mapping strategy ensures balanced verification coverage across multiple detected targets. **(4.)** The predicted positions control the camera gimbal (centering and zooming) for optical verification **(5.)** alongside laser range-finding to acquire distance and GPS coordinates. This process loops on the drone's hardware at approximately 1Hz. The magnified verification image then undergoes an offline classification to filter out outliers and identify the target **(6.)**. Finally, the identified targets are plotted on a map using their GPS coordinates **(7.)**.

We overcome these challenges by making the following contributions:

1. Our approach is inspired by the hunting strategies of birds of prey like common kestrels (*Falco tinnunculus*). When hunting, these birds hover while stabilizing their heads to detect ground-level prey motion (61,62), as illustrated in Fig. 1B and Supplementary Movie S1. To mimic this behavior, our drone hovers during inspection while rotating its camera gimbal to center and verify moving detections using optical zoom and laser-range finding for GPS mapping (Fig. 1C, Supplementary Movie S1). This scanning process runs continuously on the drone hardware (specifically, the remote controller's processor) at a frequency of around 1Hz.
2. We introduce *Temporal Max Pooling, TMP* as a simple, low-level, and lightweight signal processing framework for motion detection. By per-pixel pooling the maximal temporal value differences and integrated motion times over a substantially large (time) rolling integration window and normalizing the result, TMP transforms subtle, often imperceptible motion in individual frame-pairs into large and robustly detectable motion trails at almost no computational cost. TMP has also been field-tested in rainforest conditions and helped locate the last remaining wild Sumatran rhinos.

3. We apply *Noise-Contrastive Estimation, NCE* (63) to the TMP signal to isolate anomalous motion patterns caused by moving targets, such as animals or humans. This self-supervised learning approach requires training on only a few drone videos of empty forests recorded under varying wind conditions. The method learns the statistical distribution of background motion (primarily caused by wind, such as swaying vegetation, and drone drift) and identifies pixels that deviate from this baseline.
4. We combine the time channel from the TMP signal with Kalman filtering to predict the target's position while compensating for imaging and processing delays. These coordinates then guide the camera gimbal and laser-range finder for optical zoom verification and GPS mapping.
5. We present an algorithmic solution to ensure balanced verification coverage in case multiple targets are visible at the same time. This avoids a bias towards targets with the strongest motion signals.
6. We apply vision-based reasoning for outlier removal and classification, achieving an overall recall of 80% under realistic conditions.
7. We evaluated our approach through two field experiments – one focusing on wildlife monitoring and the other on surveillance.

Figure 2 illustrates the interaction of these steps.

# RESULTS

The *trACT* process (Fig. 2) is executed at a specified waypoint, drone heading, and gimbal pitch. Although this pose can be chosen freely, the gimbal yaw is set to zero degrees to align it with the drone heading. TMP integral images, along with their value and time encodings, are computed for this reference pose. Details on how TMP integrals are calculated are provided in Materials and Methods.

Besides the regular flight mode (F), three operating modes are supported. The manual mode (M) enables user verification of anomalous motion patterns in the value-encoding ($TMP_v$) displayed on the drone's remote control (RC). A touch selection on the RC triggers optical verification (also displayed on the RC) and GPS mapping. Further details on optical verification, mapping, and offline classification are provided in Materials and Methods.

The automatic single-target mode (AS) supports fully automatic detection and verification of a single target. If multiple targets are visible, the one with the strongest motion signal is prioritized. Details regarding motion anomaly detection and prediction are provided in Materials and Methods.

The automatic multi-target mode (AM) ensures that multiple detected targets are verified in a balanced manner without prioritizing the strongest motions. Details on multi-target verification are provided in Materials and Methods.

In addition, details regarding offline outlier removal, the classification of detected targets using vision reasoning, and the corresponding mapping procedures are provided in the Materials and Methods.

We first present the results of technical validations conducted under controlled conditions, followed by field experiments in wildlife monitoring and surveillance under challenging conditions.

## Technical Validation

Minimizing the time interval between reference-pose imaging and optical verification (achieved by centering and zooming in on detected motion) is critical, as targets may continue to move and motion prediction has inherent limitations. In our hardware implementation (see Materials and Methods), imaging requires a constant 33ms, while temporal max pooling (TMP) computations take 35ms on average (min: 20ms, max: 82ms). Motion anomaly detection averages 45ms (min: 24ms, max:123 ms), motion anomaly prediction averages 5ms (min: 0.2ms, max: 30ms), multi-target balancing averages 0.4ms (min: 0.065ms, max: 0.39ms), and optical verification averages 830ms (min: 818ms, max: 885ms). The optical verification step includes 800ms for mechanical gimbal rotation, with the remaining time allocated to zoom imaging and laser range measurements. This results in a total processing time of 946ms on average (approximately 1Hz) between initial imaging and final verification – an interval during which the target may shift position. Kalman filtering applied to the TMP time-encoding ($TMP_t$) enables the prediction of the target's location once the processing delay has elapsed, as explained in Materials and Methods. To account for runtime variations, we continuously measure and update the average processing time.

Imaging is performed using either wide-angle RGB or thermal cameras, while verification relies on the drone's RGB zoom camera operating in either daylight or night-vision mode.

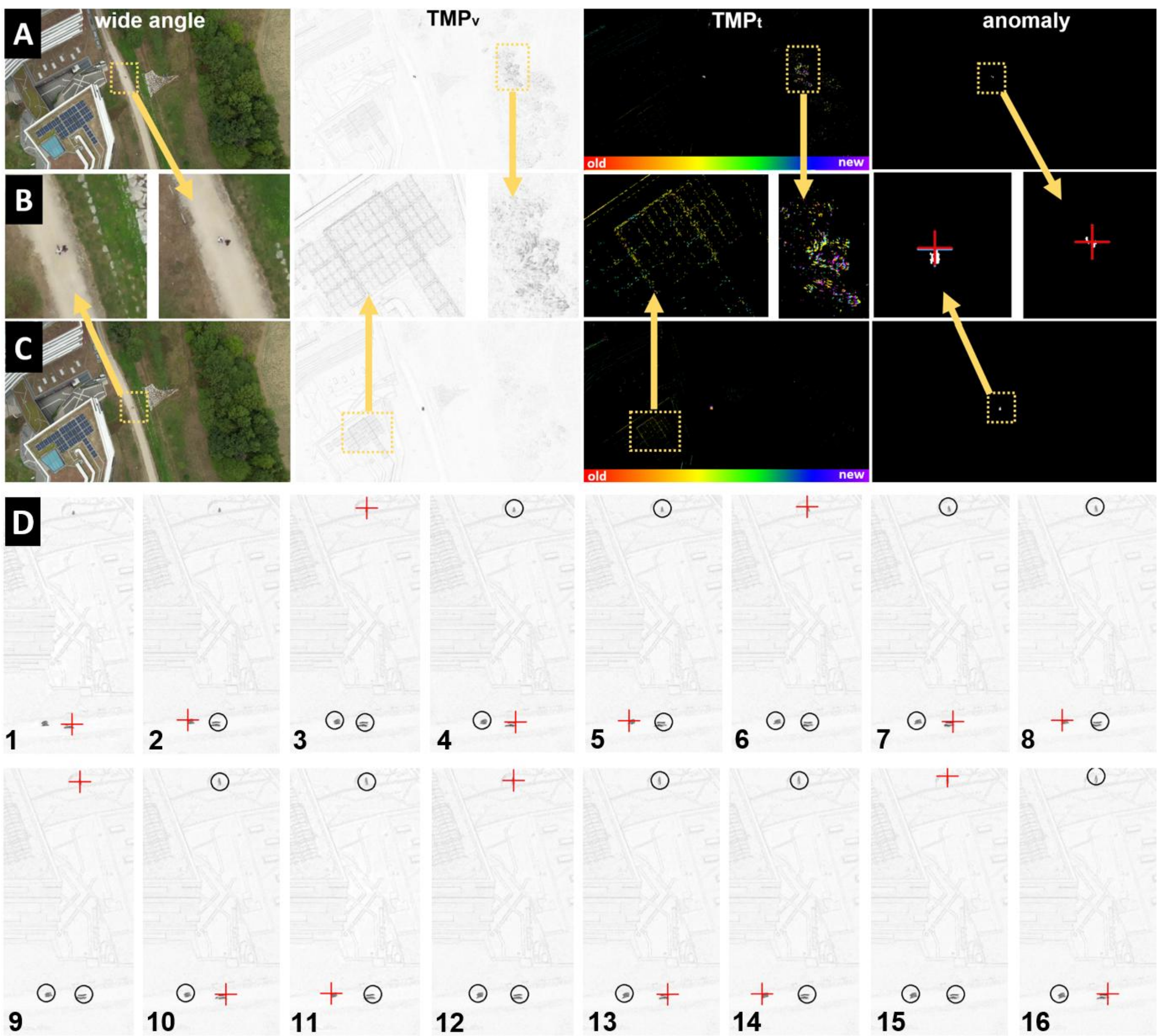

**Fig. 3. Motion anomaly detection and multi-target verification.** Wide-angle RGB experiment recorded from 120m above ground level, with background motion caused by wind gusts (row **A**) and drone drift (row **C**) clearly visible in the motion signal's value ($TMP_v$) and time ($TMP_t$) encodings. Relative time (from oldest to newest) is color-coded in $TMP_t$. Close-ups are shown in row **B**. Motion anomaly detection efficiently filters out background motion and isolates the anomalous

motion patterns caused by two walking people. Ensuring balanced verification coverage across multiple detected targets is achieved by tracking how frequently reference-pose areas (circles) have been previously verified. The example in **(D)** illustrates three moving targets across 16 time intervals, with the next target to be verified indicated by a cross. See Supplementary Movie S1.

Figure 3 illustrates a wide-angle RGB experiment recorded from an altitude of 120m above ground level (AGL) under wind speeds of 14-25 km/h. In row A, local tree motion caused by wind gusts appears in both TMP signal channels, $TMP_v$ and $TMP_t$. In row B, global environmental motion is visible in both channels due to drone drift. Row C presents close-up views. Our motion anomaly detection efficiently filters out these background signals, successfully isolating the anomalous motion trails of two walking people whose predicted positions are marked with crosses. An example of the automatic AM mode is shown in D. This mode balances verification coverage across all detected targets and avoids the bias toward the strongest motion signal as  it is the case in the AS mode. See Supplementary Movie S1.

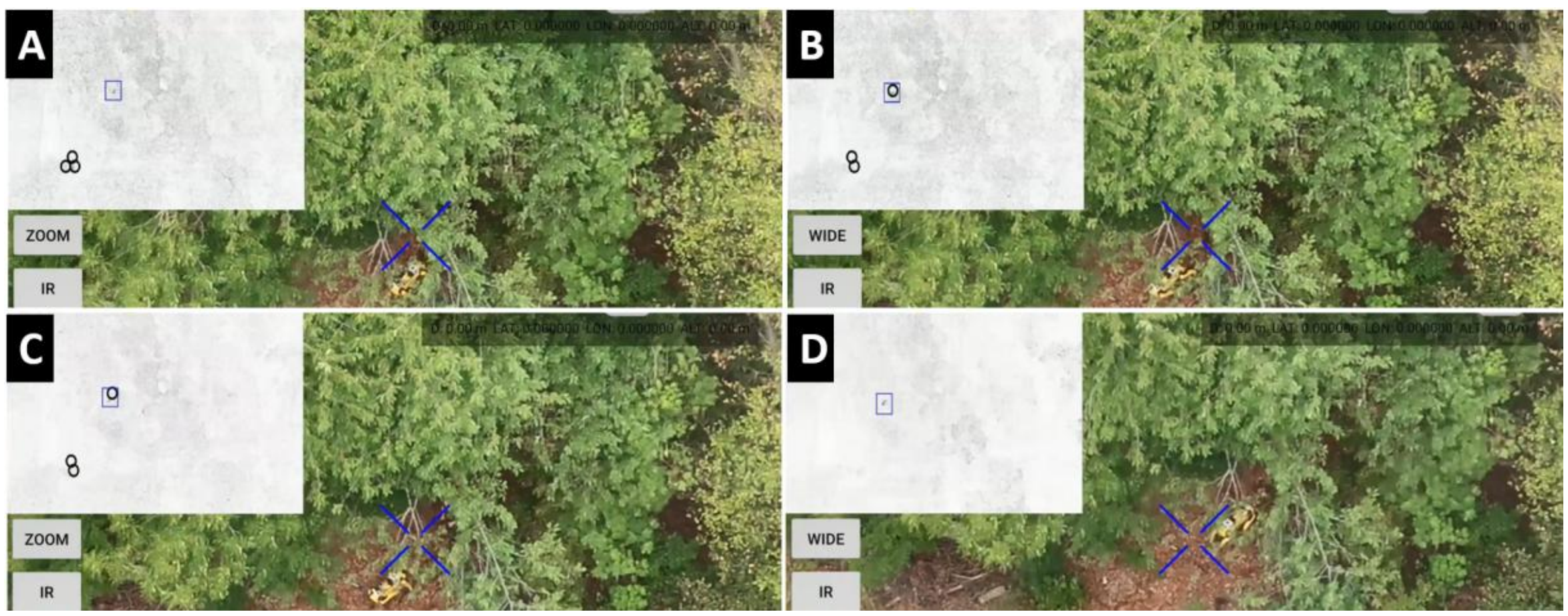

**Fig. 4. Partial occlusion.** Verification images captured from an altitude of 40m AGL under a wind speed of 14-18 km/h demonstrate the successful wide-angle RGB detection of a moving quadruped robot under partial forest occlusion. The top-left inserts display the $TMP_v$ channel highlighting the previous (circles) and current (box) verification regions. Notably, the system detected not only the robot but also nearby unoccluded people, whose verification images are omitted here. See Supplementary Movie S1.

Figure 4 shows verification images from wide-angle RGB recordings captured at an altitude of 40m AGL under a wind speed of 14-18 km/h. This experiment aimed to detect a moving, partially occluded quadruped robot in a forest environment. In AM mode, the system successfully detected not only the robot but also unoccluded people nearby. See Supplementary Movie S1.

The technical validation experiments described above were conducted under relatively controlled conditions: targets exhibited minimal camouflage and occlusion, and wind speeds were low. Because the approximate locations of the targets were known, sample video data could be recorded and processed offline to demonstrate and validate the individual processing steps. In contrast, the following sections present results from real-world field campaigns for wildlife monitoring and surveillance under more challenging conditions, featuring heavily camouflaged and hidden targets at unknown locations, with wind speeds reaching up to 40km/h. Under these operational conditions, all computations were executed live on the drone's onboard hardware.

**Field Experiment 1: Wildlife Monitoring**

Wildlife monitoring field experiments were conducted from August 28 to 31, 2026, at the Klepelshagen estate in the Uckermark region of Germany. Spanning approximately 2,500 hectares, the area is operated by the German Wildlife Foundation. Adequate monitoring locations were established by experienced personnel. The drone was manually flown to these designated waypoints to comply with visual line-of-sight regulations.

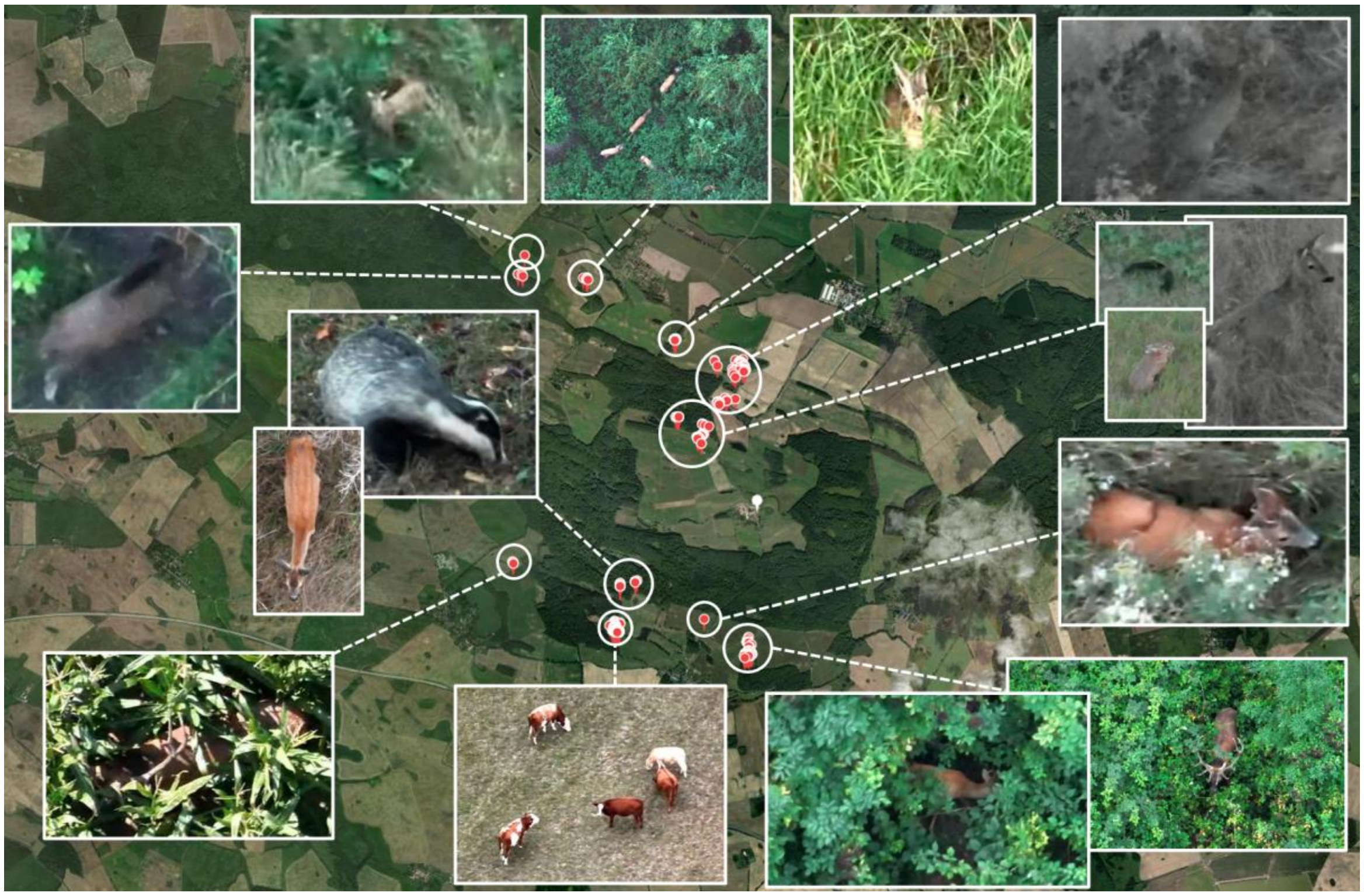

**Fig. 5. Wildlife monitoring.** Sample verification images and mapped locations of various species detected during field experiments at the Klepelshagen estate and the surrounding nature park "Am Stettiner Haff". Detections were performed both manually and automatically for single and multiple targets. See Supplementary Movies S2-S4.

A total of eight flights were conducted over a four-day experimental period (4h 26min total flight time). Flight operations spanned from dusk, through daytime, to dawn, at altitudes ranging from 60 to 120m AGL (ground resolution per pixel 8cm - 16cm) and in wind speeds between 10 and 40km/h. Wide-angle imaging was performed exclusively using the drone's thermal camera, as motion signals appear stronger in the thermal spectrum than in RGB imaging – even amid severe environmental thermal clutter. Subsequent optical verification was conducted using the drone's zoom camera (ranging from 2x to 70x magnification, depending on the flight altitude and target species) in either daylight or night-vision mode, which the camera selects automatically based on ambient light levels.

Various species were detected, ranging from small mammals, such as European hares (*Lepus europaeus*), invasive American mink (*Neogale vison*), and European badgers (*Meles meles*), to larger animals including roe deer (*Capreolus capreolus*), red deer (stags and hinds of *Cervus elaphus*), wild boar (*Sus scrofa*), and cattle (*Bos taurus*). Most of these animals were well camouflaged and/or occluded by vegetation. Figure 5 presents example verification images captured by the drone, with their corresponding detections automatically mapped. Supplementary Movie S2 demonstrates manual detection (manual mode, M). Supplementary Movie S3 illustrates automated single-target detection and verification (automatic single-target mode, AS). Supplementary Movie S4 illustrates automated multi-target detection and verification (automatic multi-target mode, AM).

In all modes, *trACT* can detect false positives if the motion detection fails. An offline classification (Fig. 2, step 6) is applied to automatically filter out false positives (i.e. verification images that do not show animals) and to classify the detected animal. Rather than relying on conventional pattern-matching approaches (64-77) – such as wildlife-fine-tuned classifiers (e.g., You Only Look Once, YOLO (78)) or dual encoders for image-text similarity (e.g., Contrastive Language-Image Pre-training, CLIP (79)) – we leverage the cognitive interpretation capabilities of advanced vision reasoning models. See Materials and Methods for details.

Throughout all field experiment flights, *trACT* captured a total of 1,276 verification images, 579 of which contained animals (recall: 46.0%). Following automatic outlier removal with vision reasoning (i.e., filtering out images without animals), the recall increased to 74.8% (precision: 99.8%, F1 score: 85.5%, accuracy: 88.3%). Furthermore, the species of the detected animals were classified with a recall of 87.6% (precision: 98.7%, F1 score: 92.8%, accuracy: 86.6%). However, a significant performance discrepancy in automatic classification was observed between daylight and night-vision images. Night-vision image classification performed worse than daylight image classification. Note that the metrics above combine daylight and night-vision verification data across all classified species. Statistics for individual flights and species are provided in the Supplementary Material (Supplementary Tables S1 and S2).

**Field Experiment 2: Surveillance**

Surveillance field experiments were conducted on September 2, 2026, at the Ruhleben Fighting City facility in Berlin – a forested training area spanning approximately 25 hectares used primarily by the Berlin Police. Volunteers stood and walked beneath the forest canopy or in open areas. To comply with visual line-of-sight regulations, the drone was manually piloted to the designated waypoints.

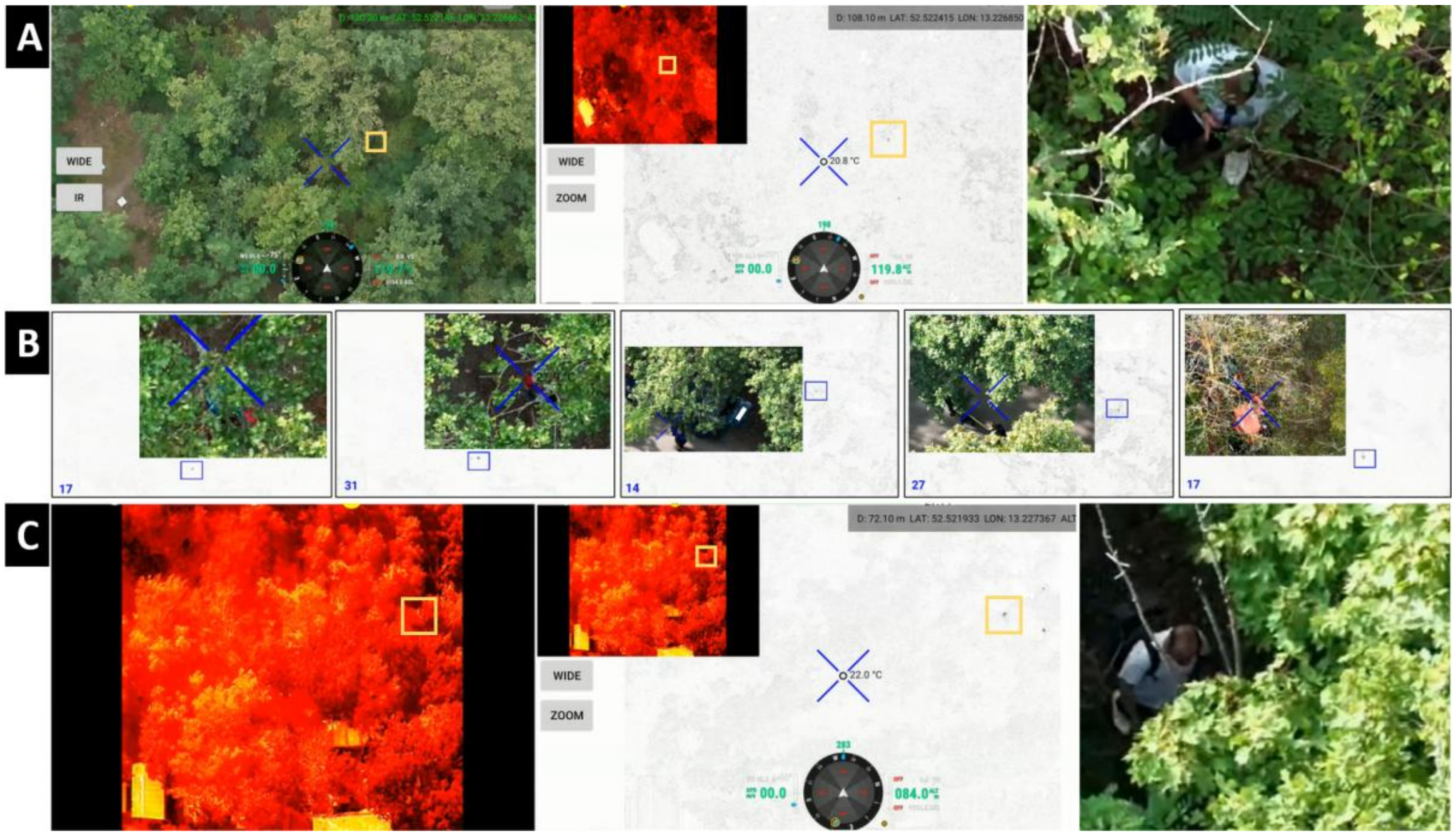

**Fig. 6. Surveillance.** Examples of walking and standing individuals detected under partial occlusion using motion signals. **(A)** A standing person revealed only by slight body movements (manual detection), with wide-angle RGB and thermal images (left and center) and the verification image (right). **(B)** Automatically detected walking people under varying degrees of occlusion, displaying the motion signals ($TMP_v$ channel) and verification images. Note that the numbers indicate the size (in pixels) of the detected motion trails. **(C)** Automatic detection of a walking person where the thermal signal is lost in background clutter, showing the wide-angle thermal image, $TMP_v$ channel, and RGB verification image. See Supplementary Movie S5.

Three flights took place between 11:00 and 17:00 local time, at altitudes ranging from 85m to 120m AGL under wind speeds between 4km/h and 25km/h. Wide-angle imaging was performed exclusively using the drone's thermal camera, resulting in typical ground resolutions from 11cm to 16cm. Subsequent optical verification in daylight mode utilized the zoom camera at magnifications from 2x to 20x, depending on flight altitude. See Supplementary Movie S5.

Across all flights, *trACT* captured 413 verification images, of which 185 contained people (yielding a recall of 44.8%). After automatic outlier removal using vision reasoning (specifically filtering out images without people) recall increased to 83.8%, achieving 100% precision, a 91.2% F1 score, and 92.7% accuracy. The mapping result of these experiments is shown in Supplementary Figure S1.

## DISCUSSION

This work introduces *trACT*, a lightweight bio-inspired robotics framework for consumer drones that bridges wide-area aerial scanning with precise, autonomous target verification. Its key contributions include Temporal Max Pooling (TMP) for motion trail extraction, self-supervised motion anomaly detection via Noise-Contrastive Estimation (NCE) to filter environmental clutter, Kalman-filtered motion prediction to compensate for system latency, and multi-target balancing for equitable verification coverage. All steps are optimized to achieve real-time detection and verification on resource-constrained edge devices, such as drone remote controllers. Under realistic field conditions during wildlife monitoring and surveillance, *trACT* achieves a raw recall of approximately 45%, which increases to 80% following vision-reasoning-based outlier removal.

Although TMP remains robust under dense partial occlusion and temperature clutter, optical verification and classification may fail when obstruction is excessive. Although occlusion-removal techniques such as synthetic aperture sensing (21) offer a potential solution, they require wide-area drone maneuvers to sample from multiple perspectives and are restricted to largely static targets. This runs counter to *trACT*'s core principle of detecting moving targets from a hovering position. Consequently, integrating these complementary sampling strategies presents a compelling direction for future research.

We found that TMP alone serves as a promising approach for reconstructing and visualizing the flight trajectories of volant animals, such as insects, bats, and birds, with initial experiments presented in Supplementary Figures S2 and S4, and Supplementary Movies S6-S9, S12, S16. Conventional methods often falter when target detection fails within individual video frames or single difference frames for background subtraction (80,81). In contrast, TMP integrates a broad temporal window of image differences, resulting in distinct and robustly detectable motion trails. Furthermore, it can be readily adapted for static ground-based cameras rather than being restricted to aerial platforms.

During an earlier wildlife conservation field expedition, we evaluated TMP in rainforest environments to detect wild Sumatran rhinos (see Supplementary Figures S3 and S4, and Supplementary Movies S10-S17). Notably, automated features such as motion anomaly detection, motion prediction, multi-target coverage, and classification were not yet implemented at that time, requiring optical verification to be performed via manual drone control. Nevertheless, TMP performed exceptionally well, particularly under conditions of extreme thermal clutter during hot daylight hours, when motion signals are nearly imperceptible in raw thermal imagery.

Employing self-supervised NCE to distinguish between anomalous motion (e.g., animals, humans) and background dynamics (e.g., wind, drone drift) is straightforward and eliminates the need for labeled training data. Furthermore, incorporating greater variability into the training set, such as varying wind speeds, flight altitudes, and target velocities, yields more robust detections. However, because the framework currently performs independent per-pixel evaluation, we believe that leveraging richer spatial and temporal context could further enhance performance. Exploring these spatiotemporal neighborhoods represents a promising direction for future research.

Of the total processing time of 946ms, 800ms is consumed by mechanical gimbal movement for optical verification. This constraint limits the system's ability to verify fast-moving targets, highlighting the need for faster gimbal mechanisms in future developments. Expanded pitch and yaw rotation ranges (e.g., those featured on the DJI Mavic 400) would allow for wider coverage without requiring drone rotation, while also eliminating the need to return the gimbal to the reference pose. Overall, this would result in smaller cumulative gimbal rotations and reduced repositioning times.

The limited flight endurance of multirotor drones remains a significant constraint, as hovering demands substantially more energy than forward flight. Future platforms leveraging the energy-efficient flight strategies of birds (82) could provide viable alternatives capable of stable, kestrel-like hovering. Finally, *trACT* is not restricted to aerial platforms, but can be readily adapted to terrestrial robotic systems or stationary mast-mounted cameras.

## MATERIALS AND METHODS

### Temporal Max Pooling

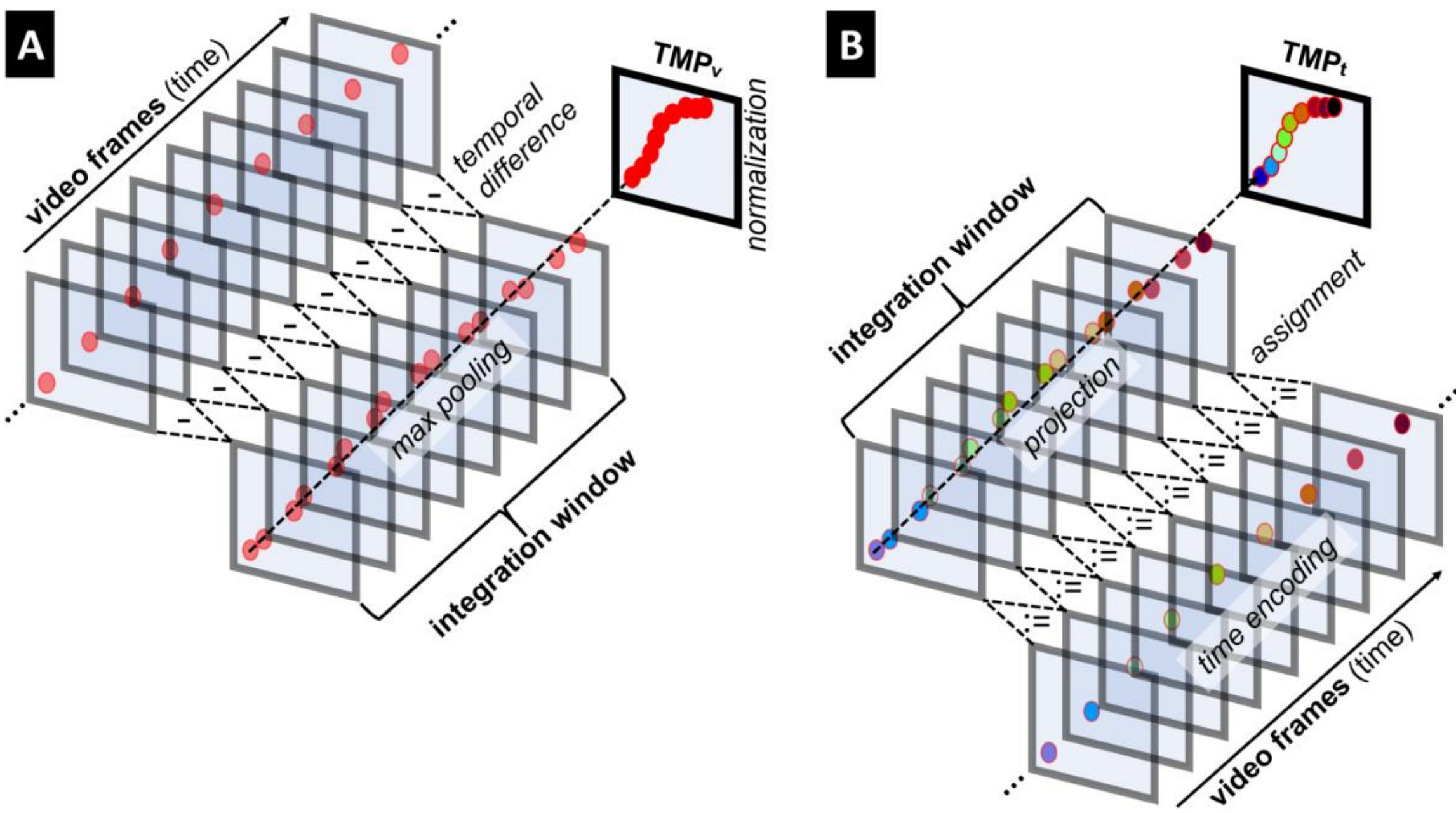

**Fig. 7. Temporal Max Pooling.** Pairwise differences of images within a fixed integration window are stacked, and the maximum values are pooled from the stack for each pixel. After normalization, this yields amplified motion trail values **(A)**. Embedding the relative time at which the maximum value is pooled per pixel produces a time encoding **(B)**. These two channels, $TMP_v$ and $TMP_t$, form our signal basis for training and inference.

Let the rolling integration window (Fig. 7A) consist of a sequence of $N$ gray scale video frames (color frames are converted to grayscale) denoted as $\{I_1, I_2, \ldots, I_N\}$, where each frame $I_k$ consist pixels at spatial coordinates $(x, y)$. First, the pairwise absolute differences between consecutive frames are computed as:

$$D_k(x,y) = |I_{k+1}(x,y) - I_k(x,y)| \quad for\ k = 1,2,\ldots,N-1.$$

Next, the stack of difference images is projected into a single image by pooling the maximum value at each pixel coordinate across time:

$$TMP(x,y) = \max_k D_k(x,y).$$

Finally, the projected image is scaled to the range [0,1] using global min-max normalization:

$$TMP_v(x,y) = \frac{TMP(x,y) - min(TMP)}{max(TMP) - min(TMP)},$$

which emphasizes the value ($TMP_v$) of the strongest motion signal within the integration window.

Alongside the magnitude of the strongest motion signal, its relative timing is also determined (Fig. 7B). Let *TS*=$\{1,2,\ldots,N\}$ represent a sequence of time steps corresponding to the order of recorded video frames within a rolling integration window. The temporal encoding ($TMP_t$) is derived by projecting the max-pooled motion signals as explained above, assigning their respective pooling time steps, and normalizing the result to the range [0,1]:

$$TMP_t(x,y) = \frac{TS(TMP_v(x,y))}{N},$$

where $TS(TMP_v)$ denotes the time step corresponding to the max-pooled value $TMP_v$. Consequently, the newest position in a motion trail corresponds to $TMP_t$=1, and the oldest corresponds to $TMP_t$=0.

In addition to the rolling integration window size $N$, a frame-skipping interval $S$ (where $S$ frames are skipped before the next frame is ingested into the window) can be defined. Assuming a constant camera frame rate $F$ (in Hz) the total integrated time window $T$ (in seconds) is given by:

$$T = \frac{(N-1)\cdot(S+1)}{F}.$$

The effective sampling rate, $R$ (in Hertz, Hz), of the resulting motion integral is:

$$R = \frac{F}{S+1}.$$

For a constant video frame rate $F$, slower motion dynamics can be captured as elongated motion trails by increasing the integration window size $N$ and/or the frame-skipping interval $S$. While increasing $S$ reduces the effective sampling rate, it improves overall processing performance and significantly mitigates memory constraints associated with storing large integration windows.

Due to hardware limitations in memory and camera frame rate, we used $N$=10 and $S$=0 for all experiments. For better visibility under direct sunlight, the live visualizations of $TMP_v$ on the drone's RC were inverted: dark regions represent high motion values, while bright regions indicate low motion.

The two signal channels ($TMP_v$ and $TMP_t$) are the basis for all following training, inference, and estimation steps.

**Motion Anomaly Detection**

To enable self-supervised learning in the absence of ground-truth motion labels, we adopted the Noise-Contrastive Estimation (NCE) strategy described in (63). NCE formulates the learning problem as a binary classification task that discriminates between observed data and artificially generated samples drawn from predefined noise distributions. The distribution of the noise must only be substantially different from the distribution of the observed data. Following this principle, we constructed pseudo-labeled training data by independently modeling the two components of each TMP channels, $TMP_v$ and $TMP_t$. As illustrated in Fig. 8, the resulting distributions of anomalous (animals) and normal (wind) motion pixels exhibit substantially different statistical characteristics, providing a basis for distinguishing the two classes in the TMP space. Specifically, the $TMP_v$ was modeled using a Gaussian distribution, whereas the $TMP_t$ was modeled using a uniform distribution. The resulting samples were assigned binary labels, with 1 denoting anomalous motion (e.g., an animal or person motion) and 0 denoting normal motion (e.g., caused by wind or drone drift). This formulation enables the motion-detection model to be trained without requiring manually annotated data.

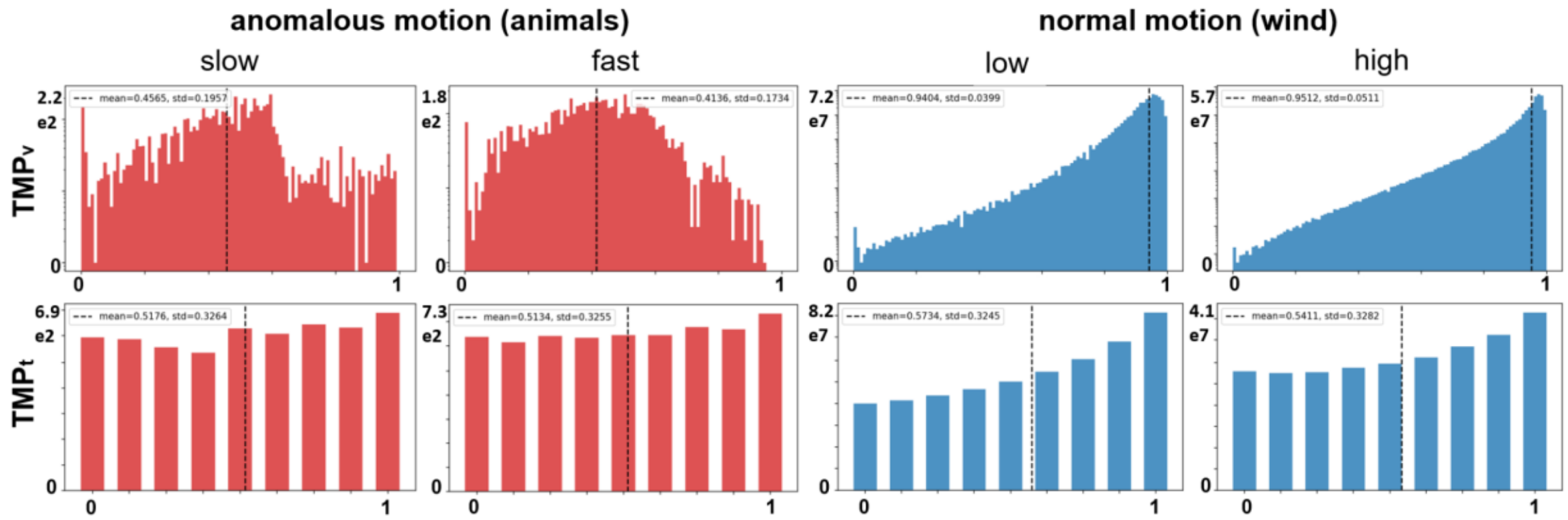

**Fig. 8. TMP distributions of anomalous motion vs normal motion.** Histograms showing pixel counts (y-axis) across TMP channels $TMP_v$ and $TMP_t$ (x-axes) for anomalous animal motion and normal wind-induced motion. Similar general distributions are observed across both high (approx. 40km/h) and low (approx. 20km/h) wind speeds, as well as between slow (standing with minor body movements) and fast (walking) animal motions. A total of 684 selected TMP images were aggregated to compute the histograms. Note that the $TMP_t$ channel is discretized to 9 relative time steps (corresponding to 9 difference images $D_k$) with $N$=10 (see Temporal Max Pooling).

Based on these pseudo-labeled samples, we employed a three-layer multilayer perceptron (MLP) to learn the distinction between normal and anomalous motion in the TMP space. Each input sample consists of the $TMP_v$ and $TMP_t$. The MLP learns a nonlinear decision boundary between the two classes and classifies each TMP according to its predicted probability of representing anomalous motion. The network was optimized using binary cross-entropy (BCE) loss.

To improve the representation of the TMP inputs, we incorporated the positional encoding function introduced in Neural Radiance Fields, NeRF (81). Positional encoding maps the input into a higher-dimensional embedding, enabling the MLP to represent high-frequency variations more effectively. The encoding function is defined as

$$\gamma(p) = (sin(2^0 \pi p), cos(2^0 \pi p), \ldots, sin(2^{L-1} \pi p), cos(2^{L-1} \pi p)),$$

where $p$ denotes an input component and $L$ is the number of frequency levels. The encoding function was applied independently to the two components of each $TMP$, i.e., the $TMP_v$ and $TMP_t$, with $L$=16.

For inference, we precomputed a lookup table containing the predicted class for every possible combination of $TMP_v$ and $TMP_t$. Specifically, the $TMP_v$ was discretized in 255 value steps, while the $TMP_t$ was discretized in 9 time steps (assuming $N$=10). This lookup-table eliminates the need for repeated forward passes and therefore enables real-time motion detection with computational requirements suitable for deployment on the drone's RC.

The MLP generates a binary prediction mask for all pixels in the input image. As a post-processing step, connected components below a predefined minimum motion trail size are removed to suppress isolated detections and small spurious responses. The minimum motion trail size is an adjustable parameter that can be configured through the drone's RC. It depends on flight altitude and target size. For each retained motion trail segment, the pixel exhibiting the maximum $TMP_t$ is identified, and its image coordinates are used to estimate the latest position of the corresponding motion.

Training and inference pipelines were implemented in Python 3.12.8 using PyTorch 2.8.0 and trained on an NVIDIA RTX 5090 GPU. Following validation of the Python implementation, the inference pipeline was reimplemented in Java to ensure compatibility with the drone's RC platform and facilitate real-time deployment.

## Motion Prediction

The $TMP_t$ channel contains the relative time steps of motion appearance per pixel. The coordinates and stored time steps of all pixels belonging to a segmented motion trail are passed to a Kalman filter (84). Combined with the average processing time (continuously measured and updated from previous iterations), the Kalman filter predicts the pixel coordinates of the motion trail for the next iteration. This coordinate is then used for centering and zoom verification. We use the *KalmanFilter* class of the OpenCV Library (85) with a constant-velocity state model $s = (x, y, v_x, v_y)^T$. Using the coordinates $(x, y)$ of all the trail pixels and their times, converted into seconds,

$$t(x, y) = N \cdot TMP_t(x, y)/R$$

derived from the $TMP_t$ channel, where $N$ is the rolling integration window size and $R$ is the effective sampling rate (see Temporal Max Pooling**),** the filter estimates the velocity $(v_x, v_y)$ of the moving target. The trail pixels are grouped by their time $t(x, y)$ and processed in temporal order. The state is initialized with the mean position of the trail pixels with the oldest $TMP_t$, i.e., at $TMP_t$=0 and zero velocity. For each group, the filter predicts the state forward and then corrects it using the positions of the pixels from that time step together. The pixel coordinates of the motion trail for the next iteration are predicted by extrapolating the estimated latest position of the corresponding motion with the estimated velocity over the average processing time.

## Optical Verification and Mapping

The predicted pixel coordinates of the motion trail are mapped to the gimbal's pitch and yaw angles to center the target on the ground after rotation. Once centered, a verification image is captured (daylight or night-vision RGB) at a pre-selected zoom level using the zoom camera. Simultaneously, a laser rangefinder measures the target's distance and calculates its ground GPS coordinates. All verification images, along with the corresponding GPS and drone coordinates, target distances, and timestamps, are stored on the remote controller. After landing, the data is transferred to a PC, where the verification images are classified and outliers are filtered out. The remaining data is then visualized in our custom, browser-based mapping application powered by OpenStreetMap.

## Classification

Rather than relying on conventional pattern-matching approaches (86-91), such as fine-tuned object detectors and classifiers for wildlife or surveillance applications (e.g., YOLO (78)) or dual-encoder models for image–text similarity (e.g., CLIP (79)), we leverage the emerging capabilities of vision-reasoning models for semantic interpretation and contextual understanding. Recent studies have demonstrated the use of vision-reasoning models for anomaly understanding in surveillance videos (92-95), wildlife species identification (96), and semantic retrieval of camera-trap imagery (97).

In our *trACT* processing framework, we employ Qwen3-VL-30B-A3B-Instruct (98) as a label-free semantic verification stage. Unlike conventional classifiers that generally require task-specific training or annotated data, the model can perform zero-shot classification through in-context prompting, allowing it to identify images containing animals and determine their corresponding species without additional fine-tuning. We therefore use the model as a post-processing step (cf. Fig. 2, step 6) to automatically filter and clean the verification dataset, retaining images that contain animals together with their predicted species (field experiment 1), as well as images containing people (field experiment 2). This process reduces the need for manual inspection and provides semantically verified samples for subsequent analysis.

For detection and species identification in our wildlife monitoring experiments, we use the following in-context prompt:

*prompt = (*
*"Look at this image carefully. Does it contain any animal? "*
*"Do not show your reasoning or thinking process. "*
*"Respond with ONLY the following two lines, nothing else:\n"*
*"Line 1: YES or NO\n"*
*"Line 2: the animal species"*
*)*

For filtering images containing people in our surveillance experiments, we use a separate prompt:

*prompt = (*
*"Look at this image carefully. Does it contain any people? "*
*"Do not show your reasoning or thinking process. "*
*"Respond with ONLY the following line, nothing else:\n"*
*"Line 1: YES or NO\n"*
*)*

Only verification images with a positive response (“Yes”) on line 1 are retained. For wildlife, the species indicated on line 2 is stored alongside the image.

**Multi-Target Verification**

We define a covermap as a list of previously verified circular regions within the reference image. Each region ($m$) is represented by its center coordinates ($m_x$, $m_y$) and a counter ($m_c$) indicating how many times the region has been previously verified. The process begins with an empty covermap list. For each imaging iteration that results in a new TMP image and the corresponding list of ($n$) detected motion trails ($t$) at coordinates ($t_x$, $t_y$) and priority $t_p = (1..n)$, the following steps are executed:

1. priority order the list of detected motion trails
2. for each motion trail $t$, determine how often its region has been verified previously:
    a. determine closest existing verification region ($m$) in the covermap
    b. if distance between ($m_x$, $m_y$) and ($t_x$, $t_y$) < RADIUS then return $m_c$, if outside any previously visited verification region the return 0
3. select motion trail(s) with lowest number of verifications, and if multiple exist with the same low number of verifications, select the one among them with the highest $t_p$
4. verify the selected motion trail at ($t_x$, $t_y$)
5. if the verified motion trail is outside any verification region in the covermap, create a new entry

$(m_x, m_y)$=$(t_x, t_y)$, $m_c$= $n$ for the closest verification region.

6. reduce $m_c$ of all $m$ by 1, and remove those entries from the covermap with $m_c$=0

A motion trail takes higher priority than another if its segmented size (post-detection) is larger. Note, that RADIUS can be adjusted. It can be derived from the verification camera's ground coverage, which depends on the field of view, zoom level, and flight altitude.

**Drone Implementation**

Our experimental results were recorded using a commercial DJI Matrice 30T drone (see Data Acquisition for details). *trACT* is deployed within a custom Android application compiled for the ARM64-based DJI RC Plus remote controller. The RC is powered by a Snapdragon 865 SoC (featuring a 1+3+4 octa-core configuration: one prime, three performance, and four efficiency cores) and runs a 64-bit Android 10 operating system. Developed using the DJI SDK (v5.18), our custom application processes live video streams received directly from the aircraft at a frame rate of *F*=30Hz. Custom controller mappings allow the user to toggle between the different modes (F, M, AS, AM), switch cameras, adjusting zoom level, choosing imaging settings (exposure, daylight and night vision), and adjust parameters (*N* (integration window size) and *S* (frame-skipping interval) for TMP, and *R* (RADIUS) and *B* (minimum motion trail size) for covermap).

**Data Acquisition**

A DJI Matrice 30T drone was used for all experiments. It is equipped with a thermal camera (uncooled VOx microbolometer with a 61° diagonal field of view, 9.1 mm focal length, f/1.0, and a resolution of 640×512 pixels at 30 Hz), a wide-angle camera (1/2-inch CMOS, 84° diagonal field of view, 4.5 mm focal length, f/2.8, and a 12 MP resolution at 30 Hz), a zoom camera (1/2-inch CMOS, 21-75 mm focal length, f/2.8-f/4.2, and a 48 MP resolution at 30Hz), and a laser rangefinder with a range of 3m to 1,200m (0.5×12m vertical area at 20% remission). The gimbal features an angular vibration range of ±0.01° and a controllable range of ±90° (pitch) and -120° to +45° (yaw). The drone was manually flown to designated waypoints to comply with visual line-of-sight regulations.

Verification images and laser-ranged GPS coordinates were stored on the remote controller and subsequently transferred to our custom desktop mapping software post-flight. Classification (including automatic outlier removal and species identification) was also performed using our custom post-flight desktop software.

**Supplementary Materials:**

Fig. S1. Mapping of surveillance experiments.
Fig. S2. Temporal Max Pooling for volant animals.
Fig. S3. Detecting a semi-wild Sumatran rhino (*Diceorhinus sumatrensis*) in thermal clutter during daylight.
Fig. S4. Additional TMP results obtained in a tropical rainforest ecosystem.
Tab. S1. Automatic outlier removal performance.
Tab. S2. Automatic species classification.
Movie S1. Airborne camera trap principle.
Movie S2. Wildlife monitoring with manual detection.
Movie S3. Wildlife monitoring with automated single-target detection and verification.

Movie S4. Wildlife monitoring with multi-target detection and verification.
Movie S5. Surveillance examples.
Software S1. *trACT* software for DJI.
Software S2. *trACT* software for Windows.
Data S1. Data for wildlife monitoring field experiments (Fig. 5).
Data S2. Data for surveillance field experiments (Fig. 6).

**Acknowledgements:** We thank the staff at Klepelshagen estate for supporting the wildlife monitoring field experiments in Germany. Drone flights at the Klepelshagen estate and surrounding areas were authorized by the District Administrator of Vorpommern-Greifswald. We also thank the staff at the German Aerospace Center (DLR) Berlin and the Berlin Police for organizing and supporting the surveillance field experiments in Berlin, Germany. Drone flights in Fighting City Berlin were authorized by the Berlin Police. All legal requirements regarding the protection of uninvolved personnel have been met. Furthermore, drone flights in Way Kambas National Park and Ujung Kulon National Park in Indonesia were authorized by the Indonesian Ministry of Forestry. We extend our sincere gratitude to all drone pilots from Yayasan Badak Indonesia (YABI / Rhino Foundation of Indonesia) and the Sumatran Rhino Sanctuary (SRS), who were trained during this campaign, for their invaluable assistance with data acquisition (drone flights and video recording).

**Funding:**
The Indonesia expedition was partially funded and co-organized by Specialized Wildlife Services Ltd, the Yayasan Badak Indonesia - the Rhino Foundation of Indonesia (YABI), and the Sumatran Rhino Sanctuary (SRS).

**Author contributions:**

Conceptualization: OB
Methodology: OB
Software: RJAAN, MY, VLB
Investigation/Implementation: OB, RJAAN, MY, VLB, RB, KH
Resources: KH, RB, OB
Visualization: OB, RJAAN, MH, VLB
Funding acquisition: OB
Project administration: OB, KH, RB
Supervision: OB
Writing – original draft: OB
Writing – review & editing: OB, RJAAN, MY, VLB, RB, KH

**Competing interests:**
Authors declare that they have no competing interests.

**Data, code, and materials availability:**
The main movies, data, and code for this study have been deposited in the database https://zenodo.org/records/23057130. Additional supplementary material on the TMP results from preceding pilot studies in rainforest environments and with volant animals has been archived in a separate database: https://zenodo.org/records/23079962. No new materials were generated in this study.

**Supplementary Materials:**

**Supplementary Materials in this file:**

Figs. S1 to S3
Tabs. S1 to S2
Legends for Movie S1 to S5
Screenshots and Legends for Software S1 to S2
Legends for Data S1 to S2

The main movies, data, and code for this study have been deposited in the database https://zenodo.org/records/23057130.

Additional supplementary material on the TMP results from preceding pilot studies in rainforest environments and with volant animals has been archived in a separate database: https://zenodo.org/records/23079962.

**Statistics for individual flights and species of wildlife monitoring experiments**

Table S1 provides individual classification statistics for flights under daylight and night-vision conditions. The accuracy of night-vision flights is notably lower than that of daylight flights. Table S2 provides classification statistics per species.

| date | flight number / type | recall | precision | F1 score | accuracy |
|---|---|---|---|---|---|
| 27/8/2026 | 1 / day light | 75.0 % | 100.0 % | 85.7 % | 95.6 % |
| 28/8/2026 | 1 / day light | 94.1 % | 100.0 % | 97.0 % | 99.0 % |
| | 2 / day light | 69.4 % | 100.0 % | 82.0 % | 87.0 % |
| | 2 / night vision | 43.6 % | 100 % | 60.7 % | 52.1 % |
| 29/8/2026 | 1 / day light | 88.9 % | 99.4 % | 93.9 % | 88.4 % |
| 30/8/2026 | 1 / day light | 97.8 % | 100 % | 98.9 % | 98.9 % |
| 31/8/2026 | 1 / day light | 96.4 % | 87.1 % | 91.5 % | 89.4 % |
| | 1 / night vision | 18.7 % | 100 % | 31.5 % | 58.9 % |
| | 2 / day light | 33.3 % | 100 % | 50.0 % | 97.0 % |
| | 3 / day light | 89.5 % | 100 % | 94.4 % | 94.4 % |

**Tab. S1. Automatic outlier removal performance.** Per-flight statistics on outlier removal (post-filtering verification images that contain animals). One to three flights were conducted per day, as indicated by the flight number. Flight type denotes whether night vision or daylight imaging was used, with the zoom camera switching automatically based on lighting conditions. Overall, performance is lower in night vision mode due to fewer image features and the absence of color saturation. A total of 1,276 verification images were captured, of which 579 contained animals (ground truth). Of these, 440 passed the outlier removal step.

| species | recall | precision | F1 score | accuracy |
|---|---|---|---|---|
| European Hares (*Lepus Europaeus*) | 70.4 % | 100 % | 82.6 % | 98.2 % |
| European Badgers (*Meles Meles*) | 83.3 % | 100 % | 90.9 % | 99.5 % |
| Wild Boar (*Sus Scrofa*) | 50.0 % | 25.0 % | 33.3 % | 99.1 % |
| Cattle (*Bos Taurus*) | 83.5 % | 100 % | 91.0 % | 96.1 % |
| Roe Deer (*Capreolus Capreolus*) / Red Deer (Stags and Hinds of *Cervus Elaphus*) | 89.5 % | 99.2 % | 94.1 % | 92.5 % |

**Tab. S2. Automatic species classification.** Per-species statistics on automatic species identification using the 440 verification images after outlier removal. Note that wild boar images were underrepresented, with only two captured. The classification model could not differentiate between deer subspecies (i.e., roe deer and red deer)

## Mapping of Surveillance Experiments

GPS locations of detected individuals (people) often reveal spatial patterns along forest paths and roads, whereas others were verified deep within the forest at designated positions.

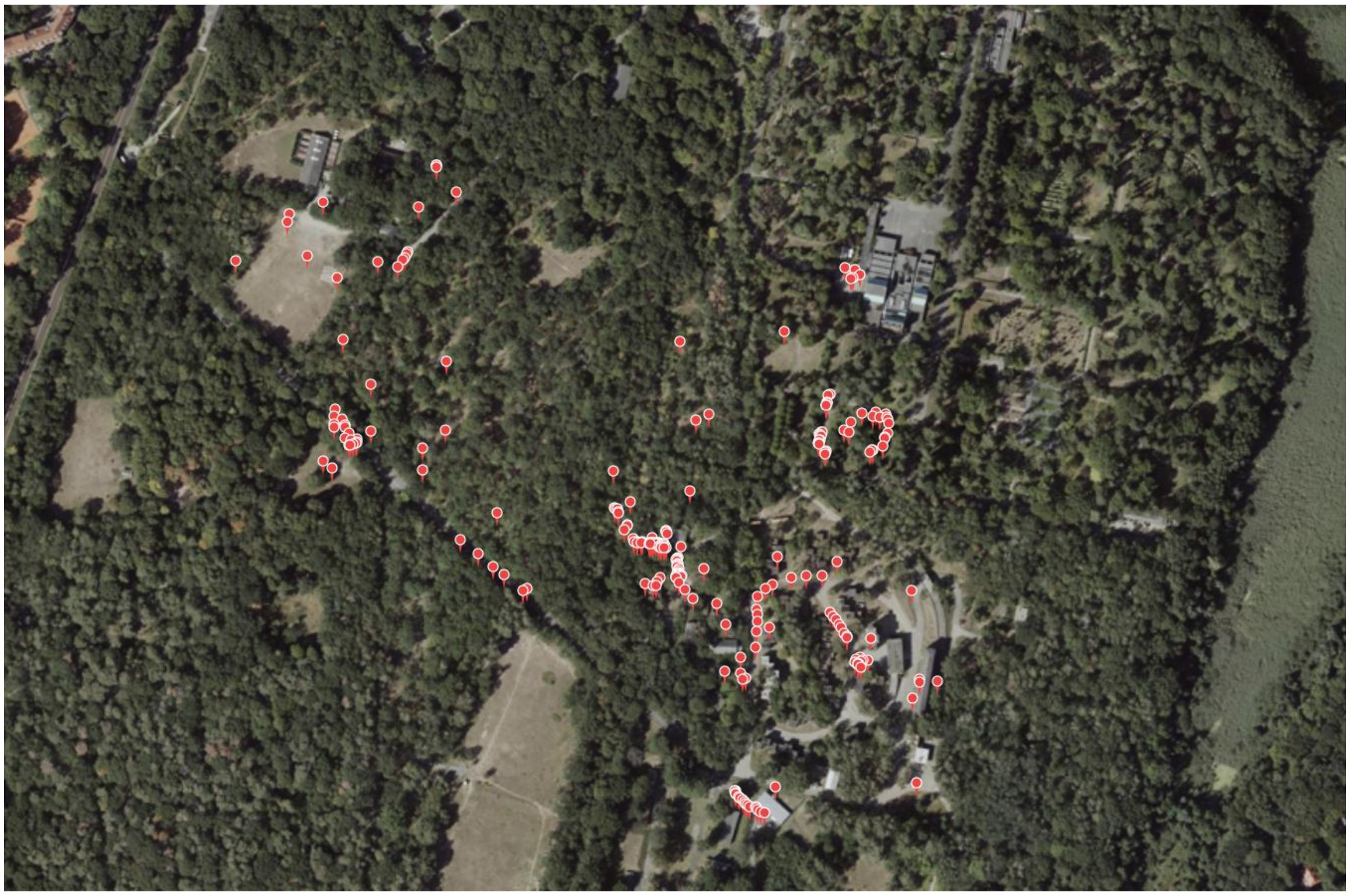

**Fig. S1. Mapping of surveillance experiments**. The satellite image displays the Ruhleben Fighting City and surrounding facilities in Berlin, alongside the mapped positions of verified targets.

## Temporal Max Pooling for Volant Animals

Applying TMP to volant animals (e.g., insects or birds) provides a straightforward way to visualize their flight paths. This approach may support future biological studies investigating such flight behavior. Because Fig. S2 presents only the TMP results without the additional *trACT* components, the corresponding video files and software have been archived in a separate database (https://zenodo.org/records/23079962).

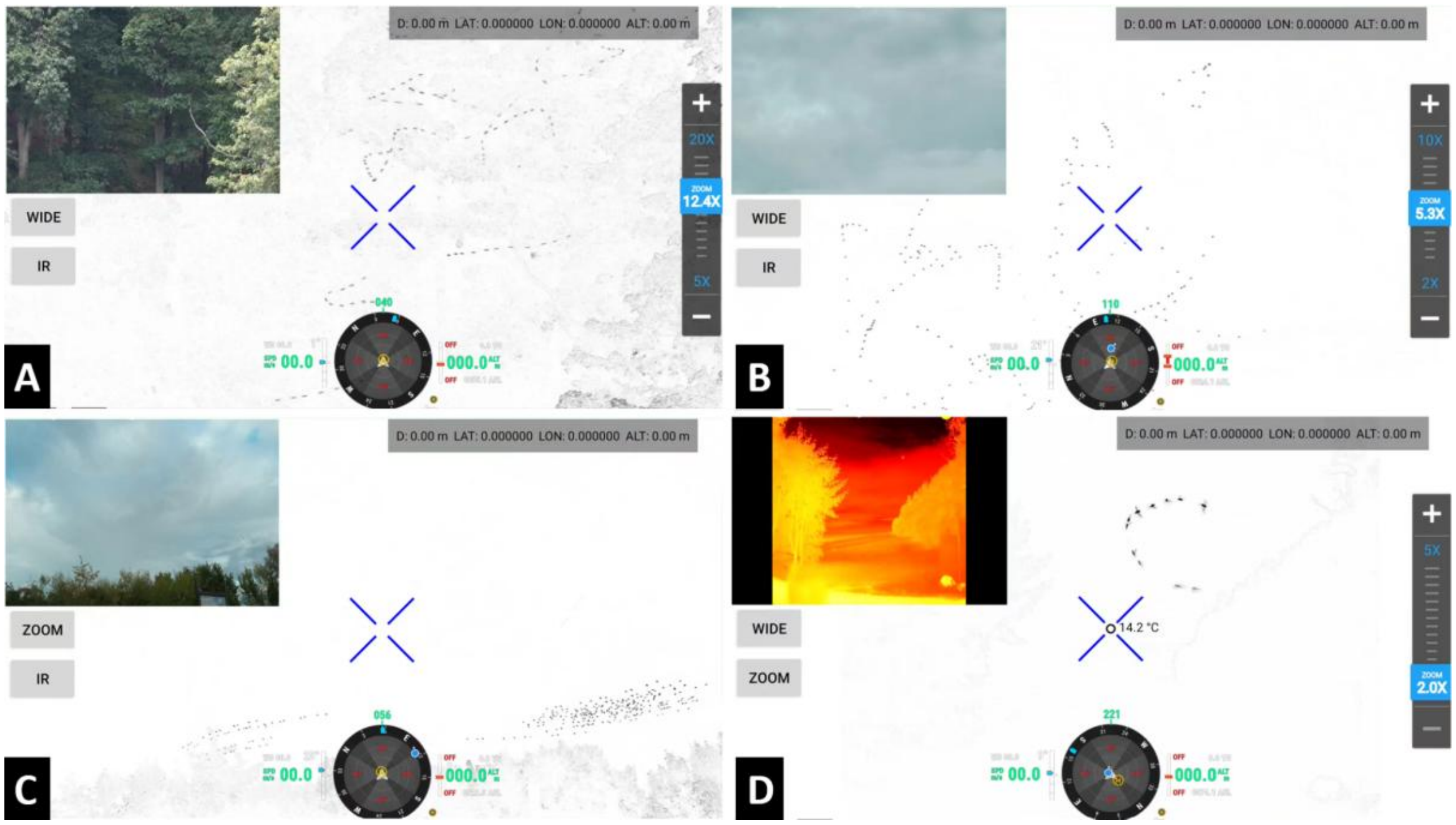

**Fig. S2. Temporal Max Pooling for volant animals.** The $TMP_v$ channel enables the visualization of flight paths. Examples shown are a swarm of dragonflies (Anisoptera) **(A)**, swarm of mosquitos (Nematocera) **(B)**, flock of European starlings (*Sturnus vulgaris*) **(C)**, and bat (Chiroptera) **(D)**. See Supplementary Movies S6-S9 (https://zenodo.org/records/23079962).

**Temporal Max Pooling Results in Rainforest Environments**

TMP (specifically only the $TMP_v$ channel) was initially used and evaluated during a field expedition in Sumatra's Way Kambas National Park (TNWK) and Java's Ujung Kulon National Park (TNUK) during May 3-14, 2026. While the primary objective of the expedition was to confirm the presence of the critically endangered wild Sumatran rhinoceros (*Diceorhinus sumatrensis*) in TNWK, our method also proved highly effective for detecting various other wildlife species and humans in tropical rainforest ecosystems (Figs. S3-S4, Supplementary Movies S10-S17, and Supplementary Software S3). Because Figs. S3-S4 depicts the findings of a pilot study preceding *trACT*, the corresponding video files and software have been archived in a separate database (https://zenodo.org/records/23079962). Note that Supplementary Software S3 computes only $TMP_v$, as used during the expedition.

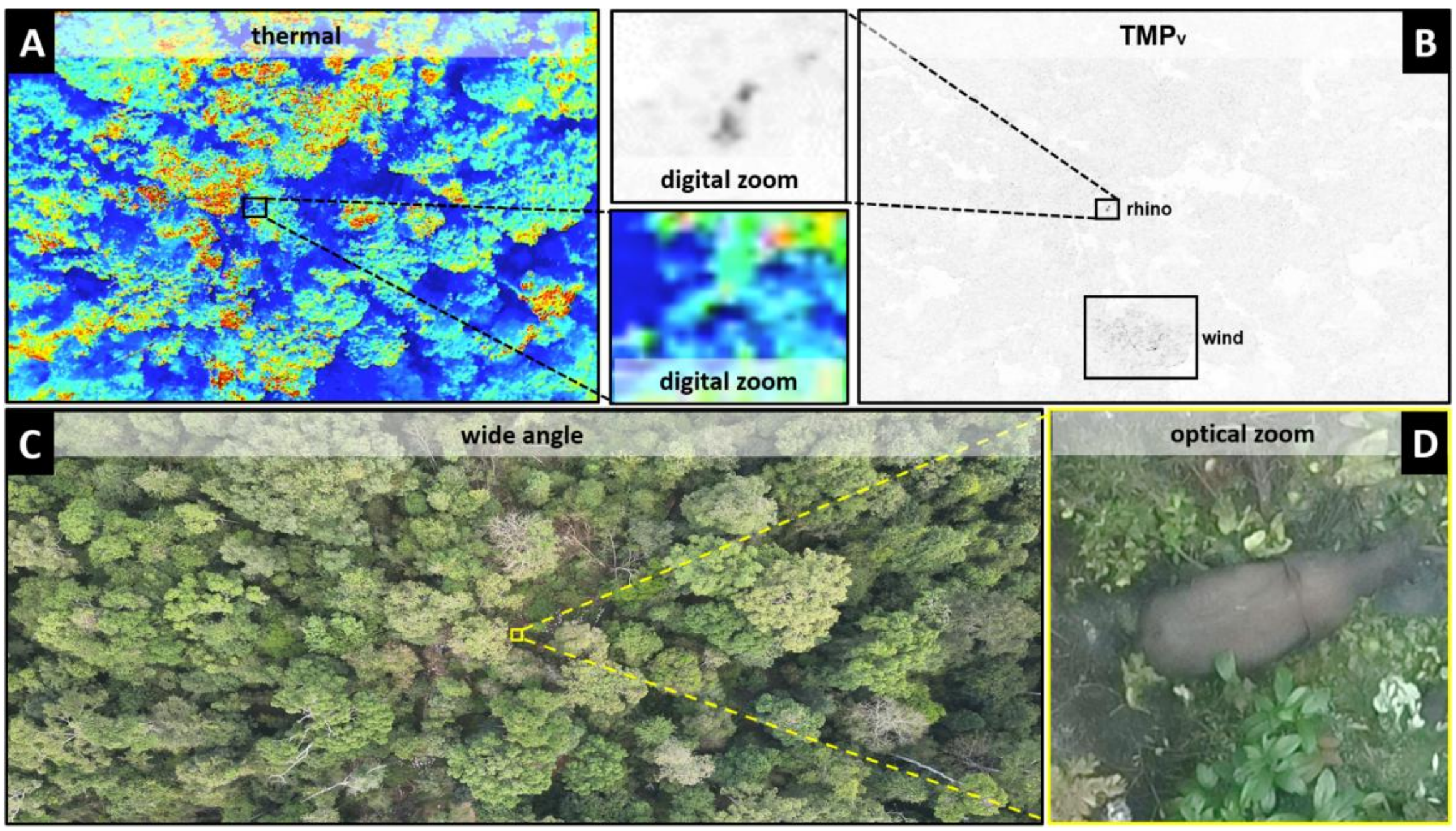

**Fig. S3. Detecting a semi-wild Sumatran rhino (*Diceorhinus sumatrensis*) in thermal clutter during daylight**. The example illustrates the detection of a rhino's motion in the $TMP_v$ channel **(B)** amid the thermal clutter of a fixed focal-length thermal camera **(A)**. The detection was achieved from an altitude of approximately 140m above ground level (AGL) and was verified using a telezoom RGB camera **(D)** while it also remained undetectable in wide-angle RGB recordings **(C)**. See Supplementary Movie S10 (https://zenodo.org/records/23079962). Note that the rhino is semi-wild and lives in a large forested paddock area in TNWK for a breeding program. It was searched and detected for training purposes. .

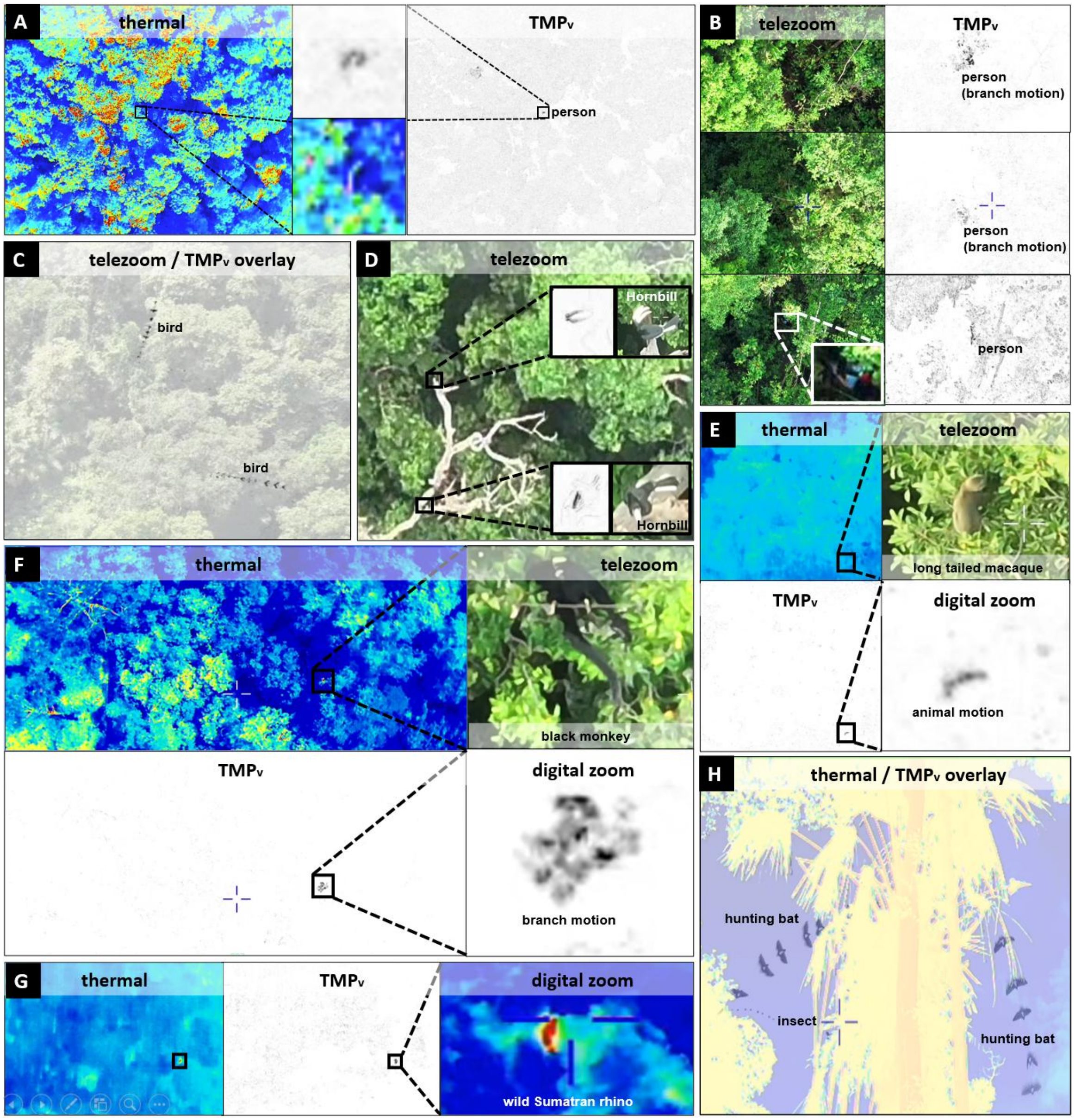

**Fig. S4. Additional TMP results obtained in a tropical rainforest ecosystem**. Shown is the $TMP_v$ channel. Body and branch motions induced by walking persons **(A,B)**, flying and perching birds **(C,D)**, smaller mammals such as long-tailed macaques (*Macaca fascicularis*) and black monkeys (*Presbytis thomasi*) **(E,F)**, flight trajectories of hunting bats and insects **(G)**, and the first visual sighting of a wild Sumatran rhino (*Diceorhinus sumatrensis)* in TNWK since 2017 **(H)**. Note that **A-F** were recorded during daylight, while **G-H** were captured at night. Acquisition altitudes ranged from 119-165m AGL. See Supplementary Movies S10-S17 (https://zenodo.org/records/23079962) corresponding to **A-H**, respectively.

Anti-poaching examples are illustrated in Figs. S4A,B, where the body motion of a walking person and the associated branch motion were detected in thermal and RGB recordings acquired from an altitude of 140-143m AGL, despite strong thermal clutter and partial occlusion (see also Supplementary Movies S10 and S11). Bird flight trajectories (captured from 165m AGL) and perching hornbills (captured from 125m AGL) were detected in RGB recordings, as shown in Figs. S4C,D (see also Supplementary Movies S12 and S13). Smaller mammals, such as a long-tailed macaque (recorded from 152m AGL, Fig. S4E and Supplementary Movie S14) and a black monkey (recorded from 119m AGL, Fig. S4F and Supplementary Movie S15), were detected in thermal recordings based either on body motion or on conspicuous branch motion. While the examples above were recorded during daylight at temperatures of up to 35°Celsius, the hunting bats and their insect prey are shown in Fig. S4G (see Supplementary Movie S16) were recorded at night from the ground. As in the case of the flying birds in Fig. S4C, TMP implicitly recovers the flight trajectories of both insects and bats. The detection of a wild Sumatran rhino (the first visual sighting in TNWK since 2017) is shown in Fig. S4H (see Supplementary Movie S17). The discovery was made at night using thermal recordings acquired from an altitude of 152m AGL. While TMP clearly detected its motion, the thermal signal alone would have been sufficient in this case, as the animal was not visually occluded and its thermal signature was clearly distinguishable from the cool ambient environment. While the shape and motion patterns in the thermal signal alone suggested that the animal was a rhinoceros, definitive confirmation was obtained through ground verification, during which fresh footprints, scratch marks, and environmental DNA (eDNA) were found.

**Movie S1. Airborne camera trap principle.** This video illustrates a hawk's fixation strategy and demonstrates its application to a drone. It first shows how *trACT* uses motion signals to find and track a herd of red deer through dense forest occlusions and thermal clutter, achieving automatic detection and verification. It then presents technical verification experiments conducted under controlled conditions, focusing on motion anomaly detection, multi-target verification, and partial occlusion.

**Movie S2. Wildlife monitoring with manual detection.** This video presents two examples of manual detection and automatic verification using motion signals from wildlife monitoring field experiments. A roe deer (*Capreolus capreolus*) was detected in dense, low-tree vegetation, and a red deer stag (*Cervus elaphus*) was located hiding in a cornfield.

**Movie S3. Wildlife monitoring with automated single-target detection and verification.** This video presents eight examples of manual detection and automatic verification using motion signals from wildlife monitoring field experiments. A red deer stag (*Cervus elaphus*) and a roe deer (*Capreolus capreolus*) were detected in dense blackberry bushes; two European hares (*Lepus europaeus*) were found hiding in tall grass and open fields; two American minks (*Neovison vison*) chasing each other in an open field; two roe deer hiding in tall grass were captured in both daylight and night-vision modes; and a European badger (*Meles meles*) was detected under a tree.

**Movie S4. Wildlife monitoring with multi-target detection and verification.** This video presents three examples of automatic detection and verification for multiple targets using motion signals. The footage features roe deer (*Capreolus capreolus*) hiding in grass in night-vision mode, and cattle (*Bos taurus*) in a pasture during daylight.

**Movie S5. Surveillance examples.** This video presents multiple examples of manual and automatic detection and verification for both unoccluded individuals and people partially occluded by forest vegetation, covering single and multiple targets.

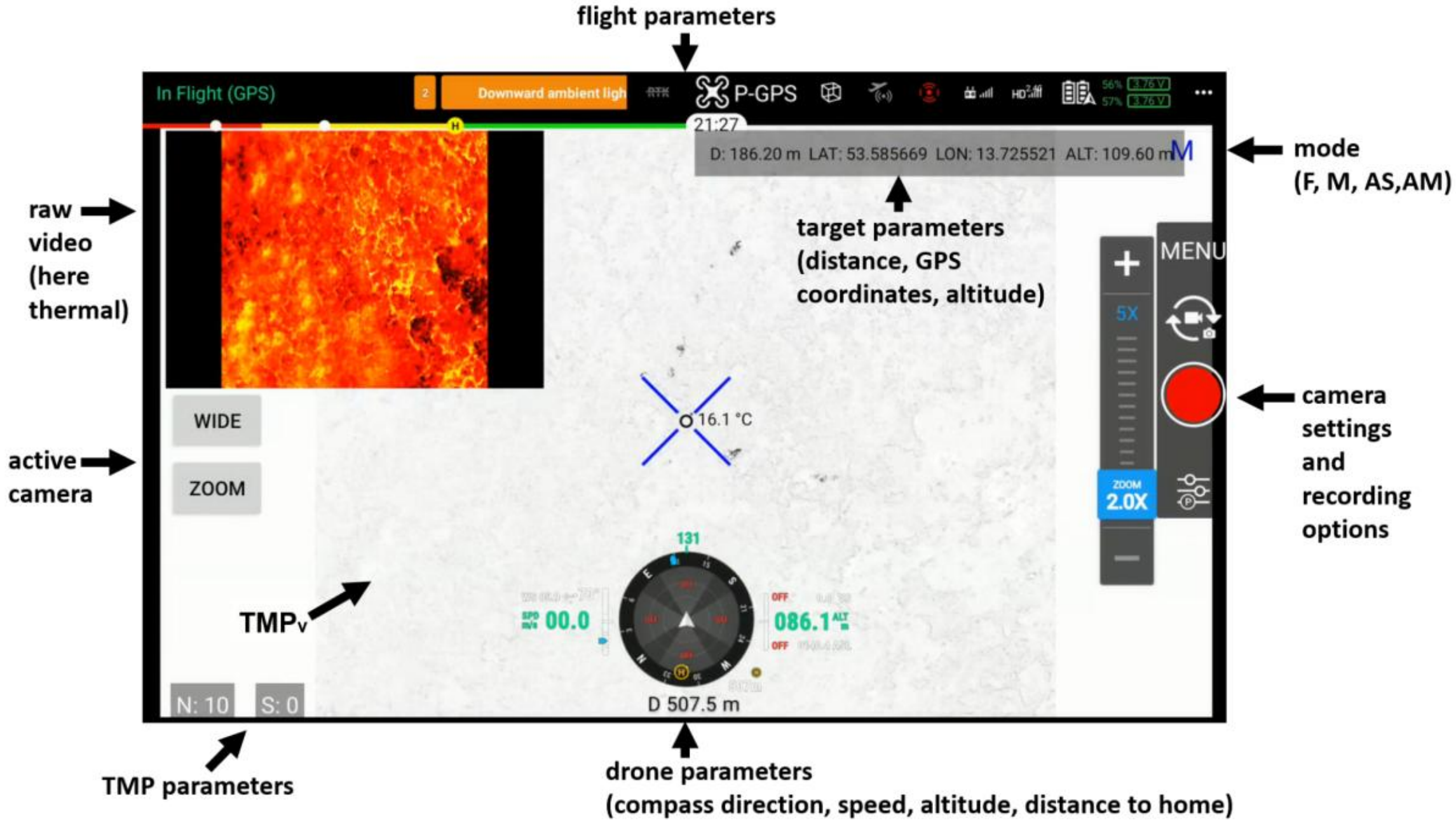

**Software S1. *trACT* software for DJI.** Screenshot of the Android application installed on the DJI RC Plus remote controller, used with the DJI Matrice 30T for all field experiments. The application supports switching models (F, M, AS, AM), configuring parameters (TMP and camera), and displaying raw video and $TMP_v$ streams alongside flight and drone telemetry. Manual detection is supported via touchscreen input on the target. Note that due to dual-use regulations, access to this application is restricted and available upon request.

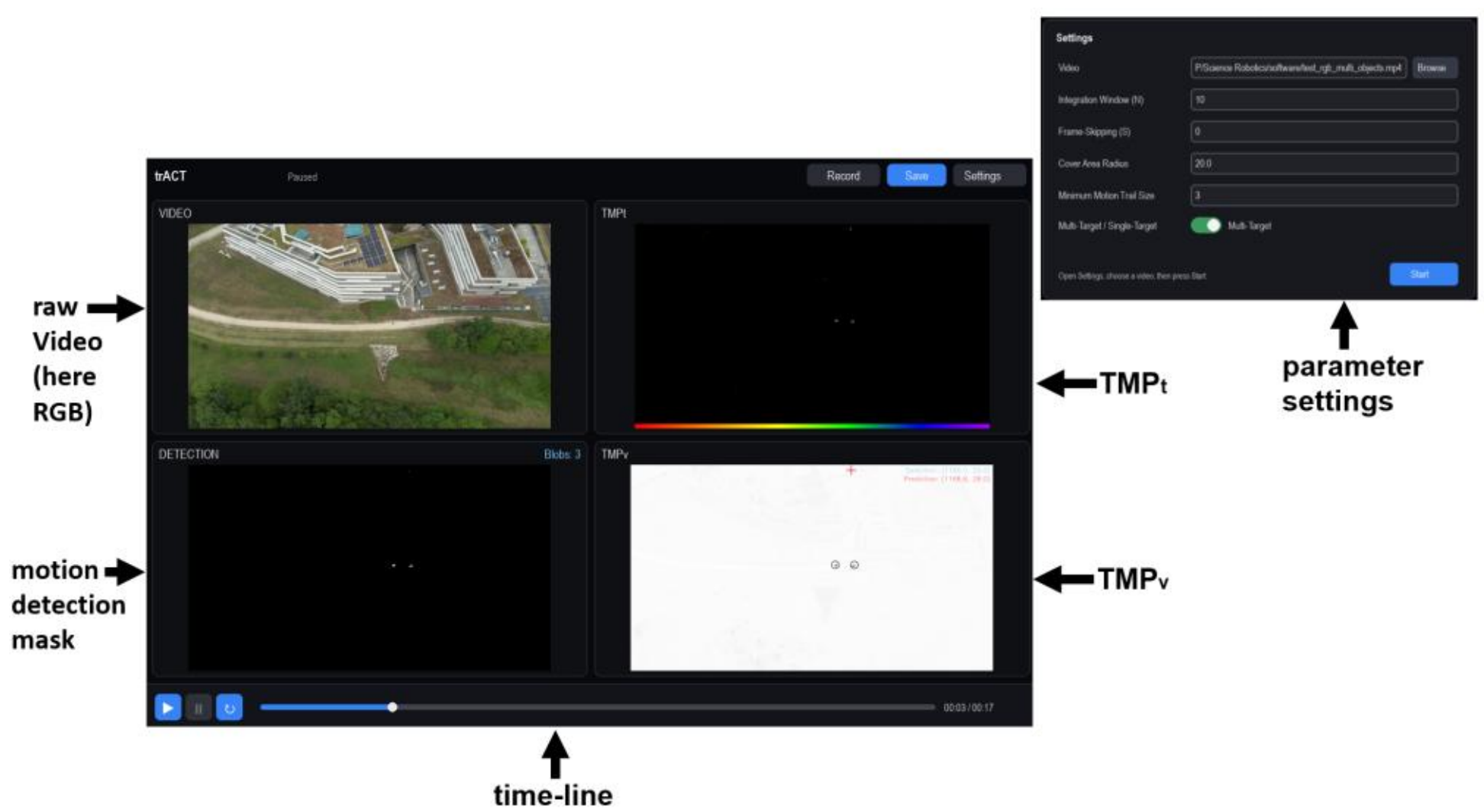

**Software S2. *trACT* software for Windows.** Screenshot of the Windows software (with Python source code available) demonstrating trACT's core functions (including TMP computations, motion detection, and multi-target verification) using pre-loaded sample videos, with support for custom video uploads.

**Data S1. Data for wildlife monitoring field experiments (Fig. 5).** The dataset contains all *trACT* verification images for the wildlife monitoring experiments (both before and after outlier removal and visual reasoning-based species classification) along with manually selected and classified ground truth images. The data are organized by flight (date and flight number, as listed in Table S1). A companion Python script for outlier removal and species classification is also provided.

**Data S2. Data for surveillance field experiments (Fig. 6).** The dataset contains all *trACT* verification images for the surveillance experiments (both before and after outlier removal) along with manually curated ground truth images. Because the flights were conducted on a single day, corresponding images are stored in the same folders. A companion Python script for outlier removal is also provided. Note that all faces have been blurred for privacy, which may prevent the reproduction of our outlier removal results.